\documentclass[11pt,a4paper,dvipsnames]{article}
\usepackage[hyperref]{acl2019}
\usepackage{times}
\usepackage{latexsym}
\usepackage{enumerate}
\usepackage{enumitem}
\usepackage{booktabs}
\usepackage{multirow}
\usepackage{tipa}
\usepackage{url}
\usepackage{xcolor}
\usepackage{amsmath}
\usepackage{amsthm}
\usepackage{amssymb}
\usepackage{newunicodechar}
\usepackage[utf8]{inputenc}
\usepackage{graphicx}
\usepackage{makecell}
\usepackage{microtype}
\usepackage{ textcomp }
\usepackage{arydshln}
\usepackage{enumitem}

\usepackage[X2,T1]{fontenc}
\usepackage{newunicodechar}
\usepackage[russian, english]{babel}
\newunicodechar{ə}{\textazerischwa}
\DeclareRobustCommand{\textazerischwa}{%
  {\fontencoding{X2}\selectfont\symbol{"BA}}%
}

\makeatletter
\expandafter\def\expandafter\@uclclist\expandafter
  {\@uclclist\textazerischwa\textazeriSchwa}
\makeatother

\usepackage{tikz}
\usetikzlibrary{calc}
\usetikzlibrary{decorations.pathmorphing}
\usetikzlibrary{shapes,arrows,chains}
\usetikzlibrary{shapes.geometric}
\usetikzlibrary{patterns}
\usetikzlibrary{positioning}
\tikzset{>=latex}
\usepackage{pgfplots}

\newcommand{\LRL}{\textsc{lrl}}
\newcommand{\HRL}{\textsc{hrl}}
\newcommand{\ENG}{\textsc{eng}}
\newcommand{\lrl}{\textsc{lrl}}
\newcommand{\hrl}{\textsc{hrl}}
\newcommand{\eng}{\textsc{eng}}
\newcommand{\umt}{\textsc{umt}}
\newcommand{\mumt}{\textsc{m-umt}}

\newcommand{\mt}{\textsc{mt}}

\newcommand{\aze}{\textsc{aze}}
\newcommand{\bel}{\textsc{bel}}
\newcommand{\glg}{\textsc{glg}}
\newcommand{\slk}{\textsc{slk}}
\newcommand{\tur}{\textsc{tur}}
\newcommand{\rus}{\textsc{rus}}
\newcommand{\por}{\textsc{por}}
\newcommand{\ces}{\textsc{ces}}

\definecolor{myyellow}{HTML}{f4e604}
\definecolor{mygreen}{HTML}{37bd79}
\definecolor{myawuamarine}{HTML}{308fac}
\definecolor{myblue}{HTML}{0457ac}

\newcommand{\data}[2]{\mathcal{#1}_{\textsc{#2}}}
\newcommand{\crdata}[3]{\hat{\mathcal{#1}}^{#2}_{\textsc{#3}}}

\newcommand{\veryshortarrow}[1][3pt]{\mathrel{%
   \hbox{\rule[\dimexpr\fontdimen22\textfont2-.2pt\relax]{#1}{.4pt}}%
   \mkern-4mu\hbox{\usefont{U}{lasy}{m}{n}\symbol{41}}}}

\makeatletter

\newcommand{\gn}[1]{\textcolor{magenta}{\bf\small [#1 --GN]}}

\newcommand{\mz}[1]{\textcolor{Blue}{\bf\small [#1 --MZ]}}
\aclfinalcopy 
\def\aclpaperid{1774} 

\newunicodechar{ə}{\textazerischwa}
\title{Generalized Data
Augmentation for Low-Resource Translation}

\author{
Mengzhou Xia, Xiang Kong, Antonios Anastasopoulos, Graham Neubig \\
  Language Technologies Institute, Carnegie Mellon University \\
  {\tt \{mengzhox, xiangk, aanastas, gneubig\}@andrew.cmu.edu} \\}

\date{}

\begin{document}
\maketitle
\begin{abstract}


Translation to or from low-resource languages (\LRL{}s) poses challenges for machine translation in terms of both adequacy and fluency. Data augmentation utilizing large amounts of monolingual data is regarded as an effective way to alleviate these problems. In this paper, we propose a general framework for data augmentation in low-resource machine translation that not only uses target-side monolingual data, but also pivots through a related high-resource language (\HRL). Specifically, we experiment with a two-step pivoting method to convert high-resource data to the \LRL, making use of available resources to better approximate the true data distribution of the \LRL. First, we inject \LRL{} words into \HRL{} sentences through an induced bilingual dictionary. Second, we further edit these modified sentences using a modified unsupervised machine translation framework. Extensive experiments on four low-resource datasets show that under extreme low-resource settings, our data augmentation techniques improve translation quality by up to~1.5 to~8 BLEU points compared to supervised back-translation baselines.%
\end{abstract}

\begin{figure}[t]
    \centering

\begin{tikzpicture}[every node/.style={minimum width=.5pt}]


\node (leg1) [draw,black,rectangle,thick,minimum height=.5pt,minimum width=.5pt] at (0,4.5) {};
\node (leg1a) [align=center,right=.1cm of leg1] {: Available Resource};
\node (leg2) [draw,black,dashed,rectangle,minimum height=.5pt,minimum width=.5pt,below=.15 of leg1] {};
\node (leg1a) [align=center,right=.1cm of leg2] {: Generated Resource};

\node (lrl1) at (1.5,0.2) [draw,black,thick,minimum height=0.4cm] {\small{LRL}};
\node (eng1) at (0.5,0.2) [draw,black,thick,minimum height=0.4cm] {\small{ENG}};
\node[] at (-0.3,0.2) {\small{[c]}};

\node (hrl1) at (1.5,1) [draw,black,thick, minimum height=0.6cm] {\small{HRL}};
\node (eng2) at (0.5,1) [draw,black,thick, minimum height=0.6cm] {\small{ENG}};
\node[] at (-0.3,1) {\small{[b]}};

\node (eng3) at (0.5,2.2) [draw,black,thick,minimum height=1.4cm] {\small{ENG}};
\node[] at (-0.3,2.2) {\small{[a]}};


\node (opt2) [draw,black,dashed,minimum height=1.4cm,below right=-.9cm and 1.35cm of eng3] {\small{HRL}};
\node (opt4) [draw,black,dashed,minimum height=1.4cm,right=1.35cm of opt2] {\small{LRL}};
\node (eng22) [draw,black,thick,minimum height=1.4cm,right=0.2cm of opt4] {\small{ENG}};

\node (opt3) [draw,black,dashed,minimum height=0.6cm,below right=0cm and 2.6cm of hrl1] {\small{LRL}};
\node (eng33) [draw,black,thick,minimum height=0.6cm,right=0.2cm of opt3] {\small{ENG}};

\node (opt1) [draw,black,dashed,minimum height=1.4cm,above=.2cm of opt4] {\small{LRL}};

\node (eng11) [draw,black,thick,minimum height=1.4cm,right=0.2 of opt1] {\small{ENG}};


\draw[->] (eng3.north) |- node[above,near end] {\tiny{[1] ENG$\rightarrow$LRL}} (opt1.west);

\draw[->] (eng3.east) |- +(0,-0.51) -- node[above,midway,text width=1cm,align=center] {\tiny{[2] ENG$\rightarrow$HRL}} (opt2.west);

\draw[->] (opt2.east) -- node[above,midway,text width=1cm,align=center] {\tiny{[4] HRL$\rightarrow$LRL}} (opt4.west);

\draw[->] (hrl1.east) -- +(.1,0) |- node[above,near end] {\tiny{[3] HRL$\rightarrow$LRL}} (opt3.west);

\end{tikzpicture}

    \caption{With a low-resource language (\LRL) and a related high-resource language (\HRL), typical data augmentation scenarios use any available parallel data [b] and [c] to back-translate English monolingual data [a] and generate parallel resources ([1] and [2]). We additionally propose scenarios [3] and [4], where we pivot through \HRL{} in order to generate a \lrl–\eng{} resource.}
    \label{fig:overview}
\end{figure}
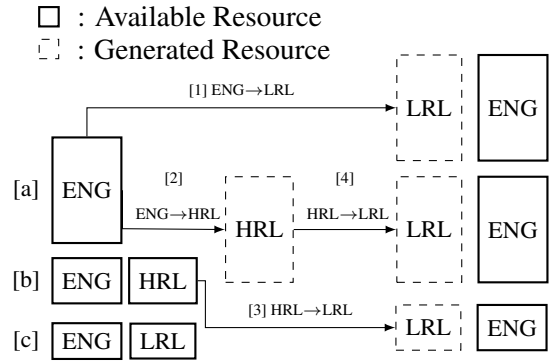

\section{Introduction}



The task of Machine Translation (MT) for low resource languages (\LRL{}s) is notoriously hard due to the lack of the large parallel corpora needed to achieve adequate performance with current Neural Machine Translation (NMT) systems \cite{koehn2017six}. A standard practice to improve training of models for an \LRL{} of interest (e.g. Azerbaijani) is utilizing data from a related high-resource language (\HRL, e.g. Turkish). Both transferring from \HRL{} to \LRL{} \cite{zoph2016transfer, nguyen2017transfer, gu2018universal} and joint training on \HRL{} and \LRL{} parallel data \cite{johnson2017google, neubig18emnlp} have shown to be effective techniques for low-resource NMT. Incorporating data from other languages can be viewed as one form data augmentation, and particularly large improvements can be expected when the \HRL{} shares vocabulary or is syntactically similar with the \LRL{} \cite{lin19acl}. Simple joint training is still not ideal, though, considering that there will still be many words and possibly even syntactic structures that will not be shared between the most highly related languages. There are model-based methods that ameliorate the problem through more expressive source-side representations conducive to sharing \cite{gu2018universal, wang2018multilingual}, but they add significant computational and implementation complexity. 

In this paper, we examine how to better share information between related \LRL{} and \HRL{}s through a framework of \emph{generalized data augmentation} for low-resource MT. In our basic setting, we have access to parallel or monolingual data of an \LRL{} of interest, its \HRL{}, and the target language, which we will assume is English. We propose methods to create pseudo-parallel \LRL{} data in this setting. As illustrated in Figure \ref{fig:overview}, we augment parallel data via two main methods: 1) back-translating from \ENG{} to \LRL{} or \HRL; 2) converting the \HRL-\eng{}  dataset to a pseudo \LRL-\ENG{} dataset.

In the first thread, we focus on creating new parallel sentences through back-translation. Back-translating from the target language to the source \cite{sennrich2015improving} is a common practice in data augmentation, but has also been shown to be less effective in low-resource settings where it is hard to train a good back-translation model \cite{currey2017copied}.
As a way to ameliorate this problem, we examine methods to instead translate from the target language to a highly-related \HRL{}, which remains unexplored in the context of low-resource NMT.
This pseudo-\HRL-\ENG{} dataset can then be used for joint training with the \LRL-\ENG{} dataset.

In the second thread, we focus on converting an \hrl-\eng{} dataset to a pseudo-\LRL{}-to-\ENG{} dataset that better approximates the true \lrl{} data. Converting between \hrl{}s and \lrl{}s also suffers from lack of resources, but because the \lrl{} and \hrl{} are related, this is an easier task that we argue can be done to some extent by simple (or unsupervised) methods.%
\footnote{This sort of pseudo-corpus creation was examined in a different context of pivoting for SMT \cite{degispert2006catalan}, but this was usually done with low-resource source-target language pairs with English as the pivot.}
In our proposed method, for the first step, we substitute \hrl{} words on the source side of \hrl{} parallel datasets with corresponding \LRL{} words from an induced bilingual dictionary generated by mapping word embedding spaces  \cite{xing-EtAl:2015:NAACL-HLT,conneau2017word}.
In the second step, we further attempt translate the pseudo-\LRL{} sentences to be closer to \LRL{} ones utilizing an unsupervised machine translation framework.



In sum, our contributions are four fold:
\begin{enumerate}[noitemsep]
    \item We conduct a thorough empirical evaluation of data augmentation methods for low-resource translation that take advantage of all accessible data, across four language pairs.
    \item
    We explore two methods for translating between related languages: word-by-word substitution using an induced dictionary, and unsupervised machine translation that further uses this word-by-word substituted data as input. These methods improve over simple unsupervised translation from \HRL{} to \LRL{} by more than~2 to~10 BLEU points.
    \item Our proposed data augmentation methods improve over standard supervised back-translation by~1.5 to~8 BLEU points, across all datasets, and an additional improvement of up to~1.1 BLEU points by augmenting from both \eng{} monolingual data, as well as \hrl-\eng{} parallel data.
\end{enumerate}


\section{A Generalized Framework for Data Augmentation}
In this section, we outline a generalized data augmentation framework for low-resource NMT.

\subsection{Datasets and Notations}
Given an \LRL{} of interest and its corresponding \HRL, with the goal of translating the \LRL{} to English, we usually have access to 1) a limited-sized \LRL-\ENG{} parallel dataset 
$\{\data{S}{le}, \data{T}{le} \}$;
2) a relatively high-resource \HRL-\ENG{} parallel dataset
$\{\data{S}{he}, \data{T}{he}\}$; 
3) a limited-sized \LRL-\HRL{} parallel dataset 
$\{ \data{S}{hl}, \data{T}{hl} \}$; 
4) large monolingual datasets in \LRL{} $\data{M}{l}$, \HRL{} $\data{M}{h}$ and English $\data{M}{e}$. 

To clarify notation, we use $\mathcal{S}$ and $\mathcal{T}$ to denote the source and target sides of parallel datasets, and $\mathcal{M}$ for monolingual data. Created data will be referred to as $\crdata{S}{m}{a$\veryshortarrow$b}$. 
The superscript $m$ denotes a particular augmentation approach (specified in Section~\ref{sec:method}).
The subscripts denote the translation direction that is used to create the data, with the \LRL, \HRL, and \ENG{} denoted with `\textsc{l}', `\textsc{h}', and `\textsc{e}' respectively.


\subsection{Augmentation from English}
The first two options for data augmentation that we explore are typical back-translation approaches:

\begin{enumerate}
\item {\bf \ENG-\LRL } We train an \ENG-\LRL{} system and back-translate English monolingual data to \LRL{}, denoted by $\{ \crdata{S}{}{e$\veryshortarrow$l}, \data{M}{e}\}$.
\item {\bf \ENG-\HRL } We train an \ENG-\HRL{} system and back-translate English monolingual data to \HRL{}, denoted by $\{ \crdata{S}{}{e$\veryshortarrow$h}, \data{M}{e}\}$.
\end{enumerate}

Since we have access to \LRL{}-\ENG{} and \HRL-\ENG{} parallel datasets,
we can train these back-translation systems \cite{sennrich2015improving} in a supervised fashion. 
The first option is the common practice for data augmentation. However, in a low-resource scenario, the created \LRL{} data can be of very low quality due to the limited size of training data, which in turn could deteriorate the \LRL$\veryshortarrow$\ENG{} translation performance. As we show in Section~\ref{sec:results}, this is indeed the case.

The second direction, using \HRL{} back-translated data for \LRL$\veryshortarrow$\ENG{} translation, has not been explored in previous work. However, we suggest that in low-resource scenarios it has  potential to be more effective than the first option because the quality of the generated \HRL{} data will be higher, and the \HRL{} is close enough to the \LRL{} that joint training of a model on both languages will likely have a positive effect.

\subsection{Augmentation via Pivoting }

Using \HRL-\ENG{} data improves \LRL-\ENG{} translation because (1) adding extra \ENG{} data improves the target-side language model, (2) it is possible to share vocabulary (or subwords) between languages, and (3) because the syntactically similar \HRL{} and \LRL{} can jointly learn parameters of the encoder. However, regardless of how close these related languages might be, there still is a mismatch between the vocabulary, and perhaps syntax, of the \HRL{} and \LRL{}.  
However, translating between \hrl{} and \lrl{} should be an easier task than translating from English, and we argue that this can be achieved by simple methods.

Hence, we propose ``Augmentation via Pivoting" where we create an \lrl-\eng{} dataset by translating the source side of \hrl-\eng{} data, into the \lrl. There are again two ways in which we can construct a new \lrl-\eng{} dataset:

\begin{enumerate}
    \setcounter{enumi}{2}
    \item \textsc{\textbf{hrl-lrl}} We assume access to an \hrl-\eng{} dataset. We then train an \hrl-\lrl{} system and convert the \hrl{} side of $\data{S}{he}$ to \lrl, creating a $\{\crdata{S}{}{h$\veryshortarrow$l}, \data{T}{he} \}$ dataset.
    \item \textsc{\textbf{eng-hrl-lrl}} Exactly as before, except that the \hrl-\eng{} dataset is the result of back-translation. That means that we have first converted English monolingual data $\data{M}{e}$ to $\crdata{S}{}{e$\veryshortarrow$h}$, and then we convert those to the \lrl{}, creating a dataset $\{\crdata{S}{}{e$\veryshortarrow$h$\veryshortarrow$l}, \data{M}{e}\}$.
\end{enumerate}

Given a \textsc{lrl-hrl} dataset $\{\data{S}{lh}, \data{T}{lh}\}$ one could also train supervised back-translation systems.
But we still face the same problem of data scarcity, leading to poor quality of the augmented datasets. Based on the fact that an \LRL{} and its corresponding \HRL{} can be similar in morphology and word order, in the following sections, we propose methods to convert \HRL{} to \LRL{} for data augmentation in a more reliable way.

\section{LRL-HRL Translation Methods}
\label{sec:method}

In this section, we introduce two methods for converting \hrl{} to \lrl{} for data augmentation. 

\subsection{Augmentation with Word Substitution}
\label{sec:3.1}
\citet{mikolov2013exploiting} show that the word embedding spaces share similar innate structure over different languages, making it possible to induce bilingual dictionaries with a limited amount of or even without parallel data \cite{xing-EtAl:2015:NAACL-HLT, zhang2017earth, conneau2017word}. 
Although the capacity of these methods is naturally constrained by the intrinsic properties of the two mapped languages, it's more likely to create a high-quality bilingual dictionary for two highly-related languages.
Given the induced dictionary, we can substitute \HRL{} words with \LRL{} ones and construct a word-by-word translated pseudo-\LRL{} corpus.



\paragraph{Dictionary Induction} We use a supervised method to obtain a bilingual dictionary between the two highly-related languages. Following \citet{xing-EtAl:2015:NAACL-HLT}, we formulate the task of finding the optimal mapping between the source and target word embedding spaces as the Procrustes problem \cite{schonemann1966generalized}, which can be solved by singular value decomposition (SVD):
{
\setlength{\abovedisplayskip}{5pt}
\setlength{\belowdisplayskip}{5pt}
\begin{gather}
    \text{$\min\limits_{W} \Vert W X - Y\Vert_F^2$ s.t.  $W^T W=I$}, \nonumber
\end{gather}
}
\noindent where $X$ and $Y$ are the source and target word embedding spaces respectively. 

As a seed dictionary to provide supervision, we simply exploit identical words from the two languages. With the learned mapping $W$, we compute the distance between mapped source and target words with the CSLS similarity measure \cite{conneau2017word}. Moreover, to ensure the quality of the dictionary, a word pair is only added to the dictionary if both words are each other's closest neighbors. Adding an \lrl{} word to the dictionary for every \hrl{} word results in relatively poor performance due to noise as shown in Section \ref{dic-ind}.

\paragraph{Corpus Construction } Given an \hrl-\eng{} $\{\data{S}{he}, \data{T}{he}\}$ or a back-translated $\{\crdata{S}{}{e$\veryshortarrow$h}, \data{M}{e}\}$ dataset, we substitute the words in $\data{S}{he}$ with the corresponding \lrl{} ones using our induced dictionary. 
Words not in the dictionary are left untouched. 
By injecting \lrl{} words, we convert the original or augmented \hrl{} data into pseudo-\lrl{}, which explicitly increases lexical overlap between the concatenated \lrl{} and \hrl{} data. The created datasets are denoted by $\{\crdata{S}{w}{h$\veryshortarrow$e}, \data{T}{he}\}$ and $\{\crdata{S}{w}{e$\veryshortarrow$h$\veryshortarrow$l}, \data{M}{e}\}$ where $w$ denotes augmentation with \textbf{w}ord substitution.  

\subsection{Augmentation with Unsupervised MT}
\label{sec:3.2}

Although we assume \lrl{} and \hrl{} to be similar with regards to word morphology and word order, the simple word-by-word augmentation process will almost certainly be insufficient to completely replicate actual \lrl{} data.
A natural next step is to further convert the pseudo-\lrl{} data into a version closer to the real \lrl{}.
In order to achieve this in our limited-resource setting, we propose to use unsupervised machine translation (\umt). 

\paragraph{UMT} Unsupervised Neural Machine Translation \cite{artetxe2017unsupervised, lample2017unsupervised, lample2018phrase} makes it possible to translate between languages without parallel data.
This is done by coupling denoising auto-encoding, iterative back-translation, and shared representations of both encoders and decoders, making it possible for the model to extend the initial naive word-to-word mapping into learning to translate longer sentences. 

Initial studies of \umt{} have focused on data-rich, morphologically simple languages like English and French. 
Applying the \umt{} framework to low-resource and morphologically rich languages is largely unexplored, with the exception of \citet{neubig18emnlp} and \citet{guzman2019two}, showing that \umt{} performs exceptionally poorly between dissimilar language pairs with BLEU scores lower than 1.
The problem is naturally harder for morphologically rich \LRL{}s due to two reasons. First, morphologically rich languages have a higher proportions of infrequent words \cite{chahuneau2013translating}.
Second, even though still larger than the respective parallel datasets, the size of monolingual datasets in these languages is much smaller compared to \HRL{}s.


\paragraph{Modified Initialization } As pointed out in \citet{lample2018phrase}, a good initialization plays a critical role in training \textsc{nmt} in an unsupervised fashion. 
Previously explored initialization methods include: 1) word-for-word translation with an induced dictionary to create synthetic sentence pairs for initial training \cite{lample2017unsupervised, artetxe2017unsupervised}; 2) joint Byte-Pair-Encoding (BPE) for both the source and target corpus sides as a pre-processing step.
While the first method intends to give a reasonable prior for parameter search, the second method simply forces the source and target languages to share the same subword vocabulary, which has been shown to be effective for translation between highly related languages. 

Inspired by these two methods, we propose a new initialization method that uses our word substitution strategy (\S\ref{sec:3.1}). Our initialization is comprised of a sequence of three steps:
\begin{enumerate}[noitemsep]
    \item First, we use an induced dictionary to substitute \hrl{} words in $\mathcal{M}_{\textsc{h}}$ to \LRL{} ones, producing a pseudo-\LRL{} monolingual dataset $\crdata{M}{}{l}$.
    \item Second, we learn a joint word segmentation model on both $\data{M}{l}$ and $\crdata{M}{}{l}$ and apply it to both datasets.
    \item Third, we train a \textsc{NMT} model in an unsupervised fashion between $\data{M}{l}$ and $\crdata{M}{}{l}$. The training objective $\mathcal{L}$ is a weighted sum of two loss terms for denoising auto-encoding and iterative back-translation:
\begin{equation}
\centering
\begin{aligned}
    \mathcal{L} =  \lambda_{1} & \big(\mathbb{E}_{x \sim \mathcal{M}_L} {- \log P_{s \veryshortarrow s} (x|C(x))} \\ 
    & + \mathbb{E}_{y \sim \hat{\mathcal{M}}_L} {- \log P_{t \veryshortarrow t} (y|C(y))}\big) \\
    + \lambda_{2} & \big(\mathbb{E}_{x \sim \mathcal{M}_L} {- \log P_{t \veryshortarrow  s} (x|u^*(y|x))}\\ 
    & + \mathbb{E}_{y \sim \hat{\mathcal{M}}_L} {- \log P_{s \veryshortarrow t} (y|u^*(x|y))}\big)
    \nonumber
\end{aligned}
\end{equation}
\noindent where $u^*$ denotes translations obtained with greedy decoding, $C$ denotes a noisy manipulation over input including dropping and swapping words randomly, $\lambda_1$ and $\lambda_2$ denotes the weight of language modeling and back translation respectively. 
\end{enumerate}

In our method, we do not use any synthetic parallel data for initialization, expecting the model to learn the mappings between a true \LRL{} distribution and a pseudo-\LRL{} distribution. This takes advantage of the fact that the pseudo-\LRL{} is naturally closer to the true \LRL{} than the \HRL{} is, as the injected \lrl{} words increase vocabulary overlap.


\paragraph{Corpus Construction} Given the word-level augmented datasets
$\{\crdata{S}{w}{h$\veryshortarrow$e}, \data{T}{he}\}$ and $\{\crdata{S}{w}{e$\veryshortarrow$h$\veryshortarrow$l}, \data{M}{e}\}$, we use the \umt{} model trained with this method to translate the pseudo-\lrl{} data from $\crdata{S}{w}{h$\veryshortarrow$e}$ and from $\crdata{S}{w}{e$\veryshortarrow$h$\veryshortarrow$l}$. We obtain new parallel datasets 
$\{\crdata{S}{m}{h$\veryshortarrow$e}, \data{T}{he}\}$ and $\{\crdata{S}{m}{e$\veryshortarrow$h$\veryshortarrow$l}, \data{M}{e}\}$ with superscript $m$ denoting \textbf{M}odified UMT (\mumt). We use superscript $u$ for un-modified standard \umt.

\subsection{Why Pivot for Back-Translation?}
\label{sec:3.3}
Pivoting through an \HRL{} in order to convert English to \LRL{} will be a better option compared to directly translating \ENG{} to \LRL{} under the following three conditions: 1) \hrl{} and \lrl{} are related enough to allow for the induction of a high-quality bilingual dictionary; 2) There exists a relatively high-resource \hrl-\eng{} dataset; 3) A high-quality \lrl-\eng{} dictionary is hard to acquire due to data scarcity or morphological distance.

Essentially, the direct \eng$\veryshortarrow$\lrl{} back-translation may suffer from both data scarcity and morphological differences between the two languages. Our proposal breaks the process into two easier steps: \eng$\veryshortarrow$\hrl{} translation is easier due to the availability of data, and \hrl$\veryshortarrow$\lrl{} translation is easier because the two languages are related.

A good example is the agglutinative language of Azerbaijiani, where each word may consist of several morphemes and each morpheme could possibly map to an English word itself. Correspondences to (also agglutinative) Turkish, however, are easier to uncover.
To give a concrete example, the Azerbijiani word ``düşünc{\textazerischwa}l{\textazerischwa}rim'' can be fairly easily aligned to the Turkish word ``düşüncelerim'' while in English it corresponds to the phrase ``my thoughts'', which is unlikely to be perfectly aligned.

\bgroup

\def\arraystretch{0.9}
\begin{table}[t]
\centering
\begin{tabular}{@{}ccccc@{}}
\toprule
        \multirow{3}{*}{Datasets} & \multicolumn{4}{c}{\lrl{} (\hrl{})}\\
         & {\aze} & {\bel} & {\glg} & {\slk} \\
         &(\tur{}) & (\rus{}) & (\por{}) & (\ces{})\\ \midrule
{$\mathcal{S}_{\textsc{le}}, \mathcal{T}_{\textsc{le}}$}  & 5.9\small{K}          & 4.5\textsc{k}         & 10\small{K}            & 61\small{K}            \\
{$\mathcal{S}_{\textsc{he}}, \mathcal{T}_{\textsc{he}}$}  & 182\small{K}        & 208\small{K}       & 185\small{K}           & 103\small{K}           \\ \midrule
{$\mathcal{S}_{\textsc{lh}}, \mathcal{T}_{\textsc{lh}}$}  & 5.7\small{K}          & 4.2\small{K}         & 3.8\small{K}             & 44\small{K}            \\ \midrule
{$\mathcal{M}_\textsc{l}$} & 2.02\small{M}       & 1.95\small{M}      & 1.98\small{M}          & 2\small{M}         \\
{$\mathcal{M}_\textsc{h}$} & 2\small{M}       & 2\small{M}     & 2\small{M}        & 2\small{M}      \\
{$\mathcal{M}_\textsc{e}$} &  \multicolumn{4}{c}  {2\textsc{m}/ 200\small{K}}        \\ \bottomrule
\end{tabular}
\caption{Statistics (number of sentences) of all datasets. }
\label{tab:datasets}
\end{table}
\egroup

\section{Experimental Setup}
\subsection{Data}
\label{datasets}
We use the multilingual TED corpus \cite{qi2018and} as a test-bed for evaluating the efficacy of each augmentation method. We conduct extensive experiments over four low-resource languages: Azerbaijani (\aze), Belarusian (\bel), Galician (\glg), and Slovak (\slk), along with their highly related languages Turkish (\tur), Russian (\rus), Portuguese (\por), and Czech (\ces) respectively.
We also have small-sized \textsc{lrl-hrl} parallel datasets, and we download Wikipedia dumps to acquire monolingual datasets for all languages.

The statistics of the parallel datasets are shown in Table \ref{tab:datasets}. For \aze{}, \bel{} and \glg{}, we use all available Wikipedia data, while for the rest of the languages we sample a similar-sized corpus. We sample 2M/200K English sentences from Wikipedia data, which are used for baseline \umt{} training and augmentation from English respectively.

\bgroup
\def\arraystretch{0.85}
\begin{table*}[t]
    \centering
    \begin{tabular}{clcrrrr}
    \toprule
    & \multicolumn{2}{l}{\multirow{3}{*}{Training Data}} & \multicolumn{4}{c}{BLEU for  \textsc{x}$\veryshortarrow$\eng{}} \\
    & & &  \multicolumn{1}{c}{\aze}  & \multicolumn{1}{c}{\bel}  & \multicolumn{1}{c}{\glg}   &  \multicolumn{1}{c}{\slk}  \\ 
    & & & (\tur{}) & (\rus{}) & (\por{}) & (\ces{}) \\
    \midrule
        \multicolumn{6}{l}{Results from Literature}\\
        & SDE \cite{wang2018multilingual} & & 12.89 & 18.71 & 31.16 & 29.16 \\
        & many-to-many \cite{aharoni2019massively} & & 12.78 & 21.73 & 30.65 & 29.54 \\
    \midrule
        \multicolumn{6}{l}{Standard \textsc{nmt}}\\
        1 & $\{\data{S}{le} \data{S}{he} \ , \  \data{T}{le} \data{T}{he}\}$  & (supervised \textsc{mt}) & 11.83 & 16.34 & 29.51 & 28.12 \\
        2 & $\{\data{M}{l}, \data{M}{e}\}$ & (unsupervised \textsc{mt}) & 0.47 & 0.18 & 1.15 & 0.75 \\
    \midrule
    
        \multicolumn{6}{l}{Standard Supervised Back-translation}\\
        3 &  \multicolumn{2}{l}{ \ + $\{\crdata{S}{s}{e$\veryshortarrow$l} \ , \  \data{M}{e}\}$} & 11.84 & 15.72 & 29.19 & 29.79 \\
        4 & \multicolumn{2}{l}{ \ + $\{\crdata{S}{s}{e$\veryshortarrow$h} \ , \  \data{M}{e}\}$} & 12.46 & 16.40 & 30.07 & 30.60 \\
    \midrule
    \multicolumn{6}{l}{Augmentation from \hrl-\eng{}}\\
        5 & \multicolumn{1}{l}{ \ + $\{\crdata{S}{s}{h$\veryshortarrow$l} \ , \  \data{T}{he}\}$} & (supervised \textsc{mt}) & 11.92 & 15.79 & 29.91 & 28.52 \\
        6 & \multicolumn{1}{l}{ \ + $\{\crdata{S}{u}{h$\veryshortarrow$l} \ , \  \data{T}{he}\}$} & (unsupervised \mt) & 11.86 & 13.83 & 29.80 & 28.69 \\
        7 & \multicolumn{1}{l}{ \ + $\{\crdata{S}{w}{h$\veryshortarrow$l} \ , \  \data{T}{he}\}$} & (word subst.) & 14.87 & 23.56 & 32.02 & 29.60 \\
        8 & \multicolumn{1}{l}{ \ + $\{\crdata{S}{m}{h$\veryshortarrow$l} \ , \  \data{T}{he}\}$} & (modified \umt) & 14.72 & 23.31 & 32.27 & 29.55 \\
        9 & \multicolumn{1}{l}{ \ + $\{\crdata{S}{w}{h$\veryshortarrow$l}\crdata{S}{m}{h$\veryshortarrow$l} \ , \  \data{T}{he}\data{T}{he}\}$} & & 15.24 & \textbf{24.25} & 32.30 & 30.00 \\
    \midrule
    \multicolumn{6}{l}{Augmention from \eng{} by pivoting}\\
        10 & \multicolumn{1}{l}{ \ + $\{\crdata{S}{w}{e$\veryshortarrow$h$\veryshortarrow$l} \ , \  \data{M}{e}\}$} & (word subst.) & 14.18 & 21.74 & 31.72 & 30.90 \\
        11 & \multicolumn{1}{l}{ \ + $\{\crdata{S}{m}{e$\veryshortarrow$h$\veryshortarrow$l} \ , \  \data{M}{e}\}$} & (modified \umt) & 13.71 & 19.94 & 31.39 & 30.22\\
    \midrule
    \multicolumn{6}{l}{Combinations}\\
    12 & \multicolumn{1}{l}{ \ + $\{\crdata{S}{w}{h$\veryshortarrow$l}\crdata{S}{w}{e$\veryshortarrow$h$\veryshortarrow$l}  \ , \  \data{T}{he}\data{M}{e}\}$} & (word subst.) & \textbf{15.74} & {\bf 24.51} & {\bf 33.16} & {\bf 32.07} \\ [.2em]
        \multirow{2}{*}{13} & \multicolumn{2}{l}{ \ + $\{\crdata{S}{w}{h$\veryshortarrow$l}\crdata{S}{m}{h$\veryshortarrow$l}\ , \  \data{T}{he}\data{T}{he}\} $} & \multirow{2}{*}{\bf 15.91} & \multirow{2}{*}{23.69} & \multirow{2}{*}{32.55}  & \multirow{2}{*}{31.58}\\
        & \multicolumn{2}{l}{ \ + $\{\crdata{S}{w}{e$\veryshortarrow$h$\veryshortarrow$l}\crdata{S}{m}{e$\veryshortarrow$h$\veryshortarrow$l} \ , \  \data{M}{e}\data{M}{e}\}$} \\
    \bottomrule
    \end{tabular}
\caption{Evaluation of translation performance over four language pairs. Rows~1 and~2 show pre-training BLEU scores. Rows~3--13 show scores after fine tuning. Statistically significantly best scores are \textbf{highlighted} ($p<0.05$).}
\label{tab:re}
\end{table*}
\egroup

\subsection{Pre-processing}
We train a joint \texttt{sentencepiece}\footnote{https://github.com/google/sentencepiece} model for each \LRL{}-\hrl{} pair by concatenating the monolingual corpora of the two languages.
The segmentation model for English is trained on English monolingual data only. We set the vocabulary size for each model to 20K. All data are then segmented by their respective segmentation model. 

We use \texttt{FastText}\footnote{https://github.com/facebookresearch/fastText} to train word embeddings using $\data{M}{l}$ and $\data{M}{h}$ with a dimension of 256 (used for the dictionary induction step). We also pre-train subword level embeddings on the segmented $\data{M}{l}$, $\crdata{M}{}{l}$ and $\data{M}{h}$ with the same dimension. 



\subsection{Model Architecture}
\label{appen:1}
\paragraph{Supervised NMT} We use the self-attention Transformer model \cite{vaswani2017attention}. We adapt the implementation from the open-source translation toolkit \texttt{OpenNMT} \cite{opennmt}. Both encoder and decoder consist of~4 layers, with the word embedding and hidden unit dimensions set to~256.  \footnote{We tuned on multiple settings to find the optimal parameters for our datasets.} We use a batch size of 8096 tokens.
\paragraph{Unsupervised NMT} We train unsupervised Transformer models with the \texttt{UnsupervisedMT} toolkit.\footnote{https://github.com/facebookresearch/UnsupervisedMT} Layer sizes and dimensions are the same as in the supervised NMT model. The parameters of the first three layers of the encoder and the decoder are shared. The embedding layers are initialized with the pre-trained subword embeddings from monolingual data. We set the weight parameters for autodenoising language modeling and iterative back translation as $\lambda_1=1$ and $\lambda_2=1$. 

\subsection{Training and Model Selection} 
After data augmentation, we follow the pre-train and fine tune paradigm for learning~\cite{zoph2016transfer, nguyen2017transfer}. We first train a base \textsc{NMT} model on the concatenation of $\{\data{S}{le}, \data{T}{le}\}$ and $\{\data{S}{he}, \data{T}{he}\}$. 
Then we adopt the mixed fine-tuning strategy of \citet{chu2017empirical}, fine-tuning the base model on the concatenation of the base and augmented datasets. For each setting, we perform a sufficient number of updates to reach convergence in terms of development perplexity.

We use the performance on the development sets (as provided by the TED corpus) as our criterion for selecting the best model, both for augmentation and final model training.



\section{Results and Analysis}
\label{sec:results}
A collection of our results with the baseline and our proposed methods is shown in Table \ref{tab:re}.

\subsection{Baselines}
The performance of the base supervised model (row 1) varies from~11.8 to~29.5 BLEU points. Generally, the more distant the source language is from English, the worse the performance. A standard unsupervised MT model (row~2) achieves extremely low scores, confirming the results of \citet{guzman2019two}, indicating the difficulties of directly translating between \lrl and \eng{} in an unsupervised fashion.
Rows~3 and~4 show that standard supervised back-translation from English at best yields very modest improvements. Notable is the exception of \slk{}-\eng{}, which has more parallel data for training than other settings. In the case of \bel{} and \glg, it even leads to worse performance.
Across all four languages, supervised back-translation into the \hrl{} helps more than into the \lrl{}; data is insufficient for training a good \lrl-\eng{} MT model. 

\subsection{Back-translation from \hrl}

\paragraph{\hrl-\LRL}
Rows 5--9  show the results when we create data using the \hrl{} side of an \hrl-\eng{} dataset. Both the low-resource supervised (row~5) and vanilla unsupervised (row~6)  \hrl$\veryshortarrow$\eng{} translation do not lead to significant improvements.
On the other hand, our simple word substitution approach (row~7) and the modified \umt{} approach (row~8) lead to improvements across the board:~+3.0 BLEU points in \aze{},~+7.8 for \bel,~+2.3 for \glg{},~+1.1 for \slk{}.
These results are significant, demonstrating that the quality of the back-translated data is indeed important. 

In addition, we find that combining the datasets produced by our word substitution and \umt{} models provide an additional improvement in all cases (row~9). Interestingly, this happens despite the fact that the \eng{} data are the exact same between rows~5--9.

\paragraph{\eng-\hrl-\LRL}
We also show that even in the absence of parallel \hrl-\lrl{} data, our pivoting method is still valuable. Rows~10 and~11 in Table~\ref{tab:re} show the translation accuracy when the augmented data are the result of our two-step pivot back-translation. In both cases, monolingual \eng{} is first translated into \hrl{} and then into \lrl{} with either just word substitution (row~10) or modified \textsc{umt} (row~11). Although these results are slightly worse than our one-step augmentation of a parallel \hrl-\lrl{} dataset, they still outperform the baseline standard back-translation (rows~3 and~4).
An interesting note is that in this setting, word substitution is clearly preferable to \umt{} for the second translation pivoting step, which we explain in \S\ref{sec:word-sent}.


\begin{figure}[t]
  \includegraphics[width=\linewidth]{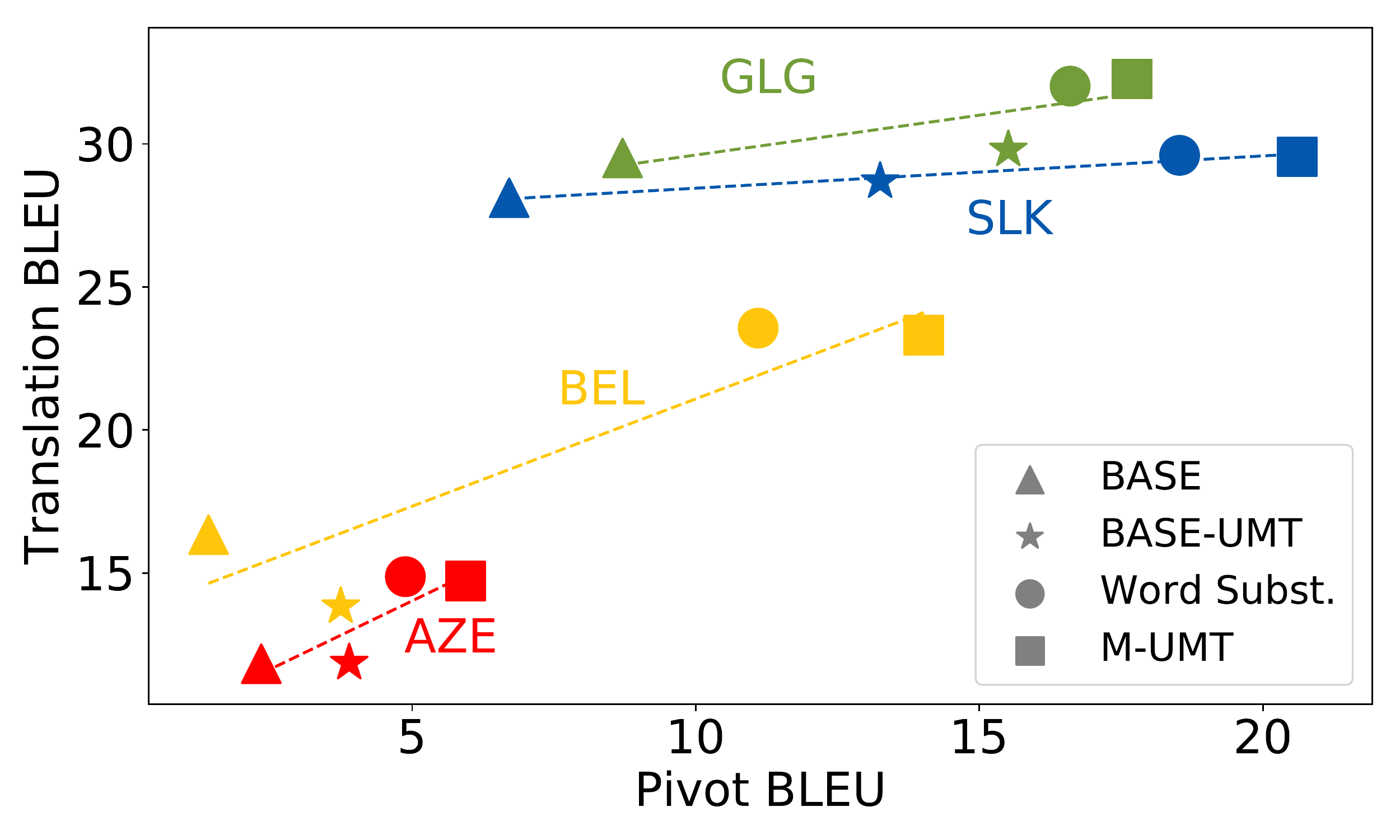}
  \caption{Correlation between \HRL{}-\lrl{} (augmentation) pivot BLEU and \LRL-\ENG{} translation BLEU.}
  \label{fig:f2}
\end{figure}
\paragraph{Combinations}
We obtain our best results by combining the two sources of data augmentation.
Row~12 shows the result of using our simple word substitution technique on the \hrl{} side of both a parallel and an artificially created (back-translated) \hrl-\eng{} dataset. In this setting, we further improve not only the encoder side of our model, as before, but we also aid the decoder's language modeling capabilities by providing \eng{} data from two distinct resources. This leads to improvements of~3.6 to~8.2 BLEU points over the base model and~0.3 to~2.1 over our best results from \hrl-\eng{} augmentation.

Finally, row~13 shows our attempt to obtain further gains by combining the datasets from both word substitution and \umt{}, as we did in setting~7. This leads to a small improvement of~0.2 BLEU points in \aze{}, but also to a slight degradation on the other three datasets. 

We also compare the results of our augmentation methods with other state-of-the-art methods that either perform improvements to modeling to improve the ability to do parameter sharing \cite{wang2018multilingual}, or train on many different target languages simultaneously \cite{aharoni2019massively}.
The results demonstrate that the simple data augmentation strategies presented here improve significantly over these previous methods.

\subsection{Analysis}
\label{sec:word-sent}

In this section we focus on the quality of \hrl$\veryshortarrow$\lrl{} translation, showing that our better \mumt{} initialization method leads to significant improvements compared to standard \umt{}.


\begin{figure}[t]
\begin{tabular}{c}
    \fcolorbox{white}{red}{\rule{0pt}{3pt}\rule{3pt}{0pt}} $\data{S}{hl}$ 
    \fcolorbox{white}{myyellow}{\rule{0pt}{3pt}\rule{3pt}{0pt}} $\crdata{S}{w}{hl}$ \
    \fcolorbox{white}{mygreen}{\rule{0pt}{3pt}\rule{3pt}{0pt}} $\crdata{S}{m}{hl}$  \
    \fcolorbox{white}{myawuamarine}{\rule{0pt}{3pt}\rule{3pt}{0pt}} $\crdata{S}{w}{hl}$+$\crdata{S}{m}{hl}$ \
    \fcolorbox{white}{myblue}{\rule{0pt}{3pt}\rule{3pt}{0pt}} $\crdata{S}{w}{hl}$+$\crdata{S}{w}{ehl}$ \\[.5em]
  \hspace{-1em}\includegraphics[width=.95\linewidth]{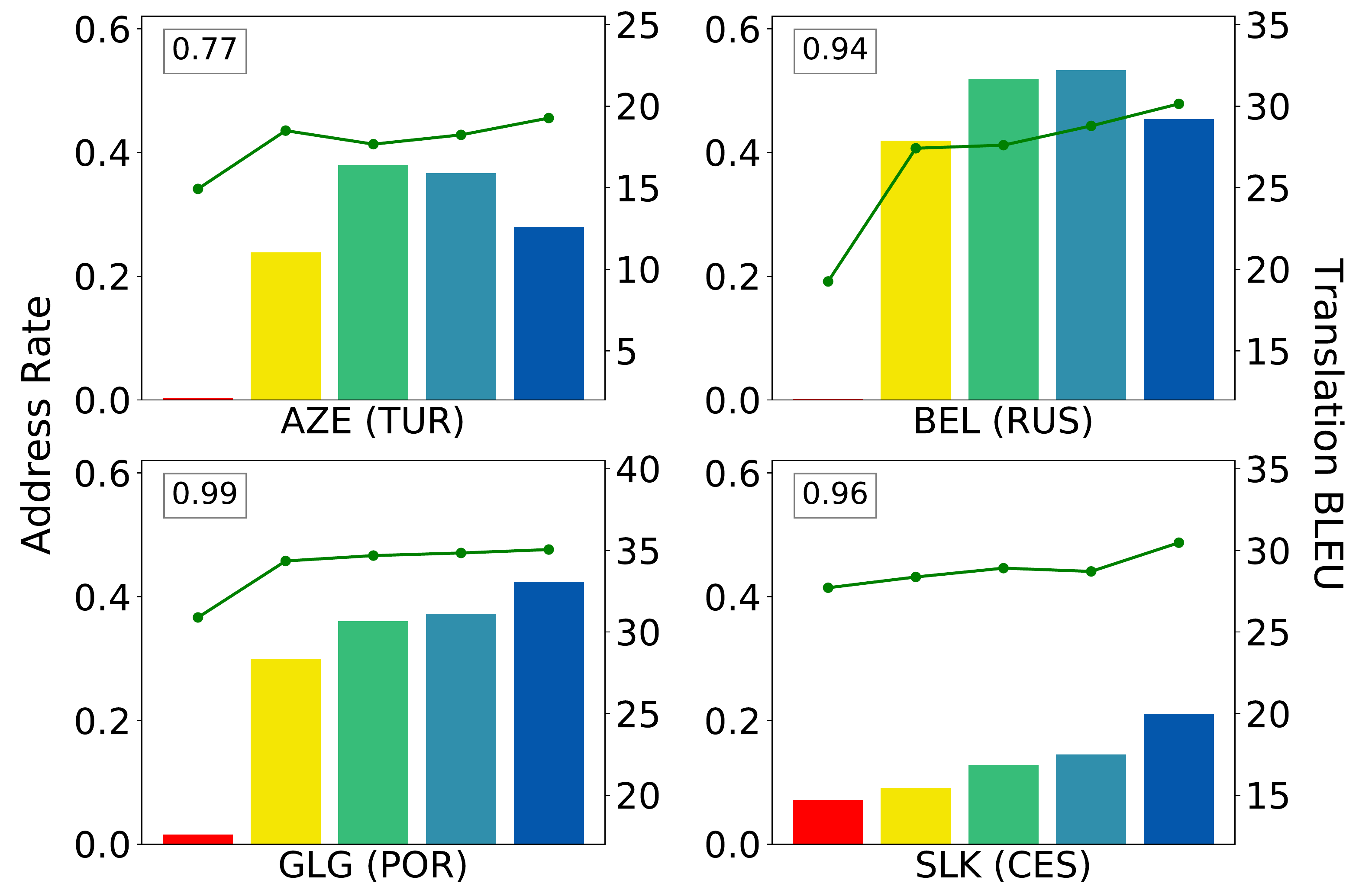}
\end{tabular}
  \caption{Rare word address rate (bars) and \lrl-\eng{} BLEU scores (line plot) for each data augmentation method. The numbers in each upper left corner is the Pearson correlation coefficient.}
  \label{fig:f1}
\end{figure}

\newcolumntype{L}{>{\centering\arraybackslash}m{0.13in}}
\newcolumntype{R}{>{\raggedright\arraybackslash}m{6.2in}}
\bgroup
\def\arraystretch{0.8}
\begin{table*}[t]
\centering
\begin{tabular}{@{}ll@{}}
\toprule
Data & \multicolumn{1}{c}{Example Sentence} \hfill Pivot BLEU\\
\midrule
$\mathcal{S}_{\textsc{le}}$ (\glg) & \small Pero con todo, veste obrigado a agardar nas mans dunha serie de estraños moi profesionais. \hfill    \\
\midrule
$\mathcal{S}_{\textsc{he}}$ (\por) & \small Em vez disso, somos obrigados a esperar nas mãos de uma série de estranhos muito profissionais.
 \hfill 0.09 \\
$\crdata{S}{w}{h$\veryshortarrow$l}$ & \small En vez disso, somos obrigados a esperar nas \textbf{mans} de \textbf{unha} \textbf{serie} de \textbf{estraños} moito \textbf{profesionais}. \hfill 0.18 \\
$\crdata{S}{m}{h$\veryshortarrow$l}$  & \small En vez diso, somos obrigados a esperar nas \textbf{mans dunha serie} de \textbf{estraños moi profesionais}. \hfill \textbf{0.54} \\
\midrule 
$\mathcal{T}_\textsc{le}$ & \small{But instead, you are forced there to wait in the hands of a series of very professional strangers.} \hfill \\
\bottomrule
\end{tabular}
\caption{A \por-\glg{} pivoting example with corresponding pivot BLEU scores. Edits by word substitution or \textsc{m-umt} are highlighted.}
\label{tab:ex_ws_umt}
\end{table*}
\egroup
We use the dev sets of the \hrl-\lrl{} datasets to examine the performance of \mumt{} between related languages.
We calculate the pivot BLEU\footnote{We will refer to \textit{pivot BLEU} in order to avoid confusion with translation BLEU scores from the previous sections.} score on the \LRL{} side of each created dataset ($\data{S}{hl}$, $\crdata{S}{w}{h$\veryshortarrow$l}$, $\crdata{S}{u}{h$\veryshortarrow$l}$, $\crdata{S}{m}{h$\veryshortarrow$l}$).
In Figure~\ref{fig:f2} we plot pivot \hrl-\lrl{} BLEU scores against the translation \lrl-\eng{} BLEU ones. 
First, we observe that across all datasets, the pivot BLEU of our \mumt{} method is higher than standard \umt{} (the squares are all further right than their corresponding stars).  Vanilla \umt{}'s scores are~2 to~10 BLEU points worse than the \mumt{} ones.
This means that \umt{} across related languages significantly benefits from initializing with our simple word substitution method.

Second, as illustrated in Figure \ref{fig:f2}, the pivot BLEU score and the translation BLEU are imperfectly correlated;
even though \mumt{} reaches the highest pivot BLEU, the resulting translation BLEU is comparable to using the simple word substitution method (rows~7 and~8 in Table \ref{tab:re}).
The reason is that the quality of $\{\crdata{S}{m}{h$\veryshortarrow$l} \ , \  \data{T}{he}\}$ is naturally restricted by the $\{\crdata{S}{w}{h$\veryshortarrow$l} \ , \  \data{T}{he}\}$, whose quality is in turn restricted by the induced dictionary.
However, by combining the augmented datasets from these two methods, we consistently improve the translation performance over using only word substitution augmentation (compare Table~\ref{tab:re} rows~7 and~9).
This suggests that the two augmented sets improve \lrl-\eng{} translation in an orthogonal way. 

Additionally, we observe that augmentation from back-translated \hrl{} data leads to generally worse results than augmentation from original \hrl{} data (compare rows~7,8 with rows~10,11 in Table~\ref{tab:re}). We believe this to be the result of noise in the back-translated \hrl{}, which is then compounded by further errors from the induced dictionary.
Therefore, we suggest that the simple word substitution method should be preferred for the second pivoting step when augmenting back-translated \hrl{} data.

Table~\ref{tab:ex_ws_umt} provides an example conversion of an \hrl{} sentence to pseudo-\LRL{} with the word substitution strategy, and its translation with \mumt{}. From $\data{S}{he}$ to $\crdata{S}{w}{h$\veryshortarrow$l}$, the word substitution strategy achieves very high unigram scores (0.50 in this case), largely narrowing the gap between two languages. The \mumt{} model then edits the pseudo-\lrl{} sentence to convert all its words to \lrl.

\paragraph{Rare Word Coverage}

\newcolumntype{L}{>{\centering\arraybackslash}m{0.35in}}
\begin{table}[t]
\centering
\begin{tabular}{lLLLL}
\toprule
         & \aze & \bel & \glg  & \slk \\ 
         & (\tur) & (\rus) & (\por) & (\ces) \\ \midrule
WT-Bi & 35\textsc{K}   & 42\textsc{K}   & 34\textsc{K}   & 51\textsc{K}   \\
WT-Uni    & 211\textsc{K}   & 179\textsc{K}   & 89\textsc{K}   & 117\textsc{K}   \\
\midrule
WN-Bi & 1.6\textsc{M}   & 2.5\textsc{M}   & 3.1\textsc{M}   & 2.0\textsc{M}  \\
WN-Uni    & 2.9\textsc{M}   & 3.8\textsc{M}   & 3.8\textsc{M}   & 2.9\textsc{M}   \\
\midrule
BLEU-Bi & 14.33   & 21.55   & 31.72   & 29.09   \\
BLEU-Uni    & 14.10   & 21.86   & 30.51   & 28.58   \\
\bottomrule
\end{tabular}
\caption{Injected word type (WT), injected word number (WN) and BLEU score (BLEU) on low-resource translation with different induced dictionaries. \textbf{Bi} denotes \textbf{bi}directional and \textbf{Uni} denotes \textbf{uni}directional word induction.}
\label{tab:biuni}
\end{table}
Next, we quantitatively evaluate how our pivoting augmentation methods increase rare word coverage and the correlation with \lrl-\eng{} translation quality. For each word in the tested set, we define a word as ``rare'' if it is in the training set's lowest 10\textsuperscript{th} frequency percentile. This is particularly true for \LRL{} test set words when using concatenated \HRL{}-\LRL{} training data, as the \LRL{} data will be smaller. We further define rare words to be ``addressed'' if after adding augmented data the rare word is not in the lowest 10\textsuperscript{th} frequency percentile anymore. Then, we define the ``address rate'' of a test dataset as the ratio of the number of addressed words to the number of rare words.
The address rate of each method, along with the corresponding translation BLEU score is shown in Figure~\ref{fig:f1}. As indicated by the Pearson correlation coefficients, these two metrics are highly correlated, indicating that our augmentation methods significantly mitigate problems caused by rare words, improving MT quality as a result.

\paragraph{Dictionary Induction}
\label{dic-ind}
We conduct experiments to compare two methods of dictionary induction from the mapped word embedding spaces: 1) Unidirectional: For each \hrl{} word, we collect its closest \lrl{} word to be added to the dictionary; 2) Bidirectional: We only add word pairs the two words of which are each other's closest neighbor to the dictionary.

In order to know how many \lrl{} words are injected into the \hrl{} corpus, we show the number of injected unique word types, number of injected words, and the corresponding BLEU score of models trained with bidirectional and unidirectional word induction in Table~\ref{tab:biuni}. It can be seen that the ratio of word numbers is higher than that of word types between bidirectional and unidirectional word induction, indicating that the injected words using the bidirectional method are of relatively high frequency.  The BLEU scores show that bidirectional word induction performs better than unidirectional induction in most cases (except \bel). One explanation could be that adding each word's closest neighbor as a pair into the dictionary introduces additional noise that might harm the low-resource translation to some extent.

\section{Related Work}
Our work is related to multilingual and unsupervised translation, bilingual dictionary induction, as well as approaches for triangulation (pivoting).

In a low-resource MT scenario, multilingual training that aims at sharing parameters by leveraging parallel datasets of multiple languages is a common practice. Some works target learning a universal representation for all languages either by leveraging semantic sharing between mapped word embeddings \cite{gu2018universal} or by using character n-gram embeddings \cite{wang2018multilingual} optimizing subword sharing. More related with data augmentation, \citet{nishimura2018multi-b} fill in missing data with a multi-source setting to boost multilingual translation. 

Unsupervised machine translation enables training NMT models without parallel data \cite{artetxe2017unsupervised, lample2017unsupervised,lample2018phrase}. Recently, multiple methods have been proposed to further improve the framework. By incorporating a statistical MT system as posterior regularization, \citet{ren2019unsupervised} achieved state-of-the-art for en-fr and en-de MT. Besides MT, the framework has also been applied to other unsupervised tasks like non-parallel style transfer \cite{lample2018styletransfer, zhang2018style}. 

Bilingual dictionaries learned in both supervised and unsupervised ways have been used in low-resource settings for tasks such as named entity recognition \cite{xie2018neural} or information retrieval \cite{litschko2018unsupervised}. \citet{hassan2017synthetic} synthesized data with word embeddings for spoken dialect translation, with a process that requires a \textsc{lrl-eng} as well as a \textsc{hrl-lrl} dictionary, while our work only uses a \textsc{hrl-lrl} dictionary.

Bridging source and target languages through a pivot language was originally proposed for phrase-based MT \cite{degispert2006catalan,cohn2007machine}. It was later adapted for Neural MT \cite{levinboim-chiang:2015:EMNLP}, and \citet{cheng2017joint} proposed joint training for pivot-based NMT. \citet{chen2017teacher} proposed to use an existing pivot-target NMT model to guide the training of source-target model. \citet{lakew2018improving} proposed an iterative procedure to realize zero-shot translation by pivoting on a third language.

\section{Conclusion}
We propose a generalized data augmentation framework for low-resource translation, making best use of all available resources. We propose an effective two-step pivoting augmentation method to convert \HRL{} parallel data to \LRL. In future work, we will explore methods for controlling the induced dictionary quality to improve word substitution as well as \mumt. We will also attempt to create an end-to-end framework by jointly training \mumt{} pivoting system and low-resource translation system in an iterative fashion in order to leverage more versions of augmented data.

\section*{Acknowledgements}
The authors thank Junjie Hu and Xinyi Wang for discussions  on the paper. This material is based upon work supported in part by the Defense Advanced Research Projects Agency Information Innovation Office (I2O) Low Resource
Languages for Emergent Incidents (LORELEI)
program under Contract No. HR0011-15-C0114 and the National Science Foundation under grant 1761548.
The views and conclusions contained in this document are those of the authors and should not be
interpreted as representing the official policies, either expressed or implied, of the U.S. Government. The U.S. Government is authorized to reproduce and distribute reprints for Government
purposes notwithstanding any copyright notation
here on.

\bibliography{References}

\begin{thebibliography}{38}
\expandafter\ifx\csname natexlab\endcsname\relax\def\natexlab#1{#1}\fi

\bibitem[{Aharoni et~al.(2019)Aharoni, Johnson, and
  Firat}]{aharoni2019massively}
Roee Aharoni, Melvin Johnson, and Orhan Firat. 2019.
\newblock Massively multilingual neural machine translation.
\newblock In \emph{Proceedings of the 2019 Conference of the North American
  Chapter of the Association for Computational Linguistics: Human Language
  Technologies}.

\bibitem[{Artetxe et~al.(2018)Artetxe, Labaka, Agirre, and
  Cho}]{artetxe2017unsupervised}
Mikel Artetxe, Gorka Labaka, Eneko Agirre, and Kyunghyun Cho. 2018.
\newblock Unsupervised neural machine translation.
\newblock In \emph{International Conference on Learning Representations}.

\bibitem[{Chahuneau et~al.(2013)Chahuneau, Schlinger, Smith, and
  Dyer}]{chahuneau2013translating}
Victor Chahuneau, Eva Schlinger, Noah~A Smith, and Chris Dyer. 2013.
\newblock Translating into morphologically rich languages with synthetic
  phrases.
\newblock In \emph{Proceedings of the 2013 Conference on Empirical Methods in
  Natural Language Processing}, pages 1677--1687.

\bibitem[{Chen et~al.(2017)Chen, Liu, Cheng, and Li}]{chen2017teacher}
Yun Chen, Yang Liu, Yong Cheng, and Victor~OK Li. 2017.
\newblock A teacher-student framework for zero-resource neural machine
  translation.
\newblock In \emph{Proceedings of the 55th Annual Meeting of the Association
  for Computational Linguistics (Volume 1: Long Papers)}, pages 1925--1935.

\bibitem[{Cheng et~al.(2017)Cheng, Yang, Liu, Sun, and Xu}]{cheng2017joint}
Yong Cheng, Qian Yang, Yang Liu, Maosong Sun, and Wei Xu. 2017.
\newblock Joint training for pivot-based neural machine translation.
\newblock In \emph{Proceedings of the 26th International Joint Conference on
  Artificial Intelligence}, pages 3974--3980.

\bibitem[{Chu et~al.(2017)Chu, Dabre, and Kurohashi}]{chu2017empirical}
Chenhui Chu, Raj Dabre, and Sadao Kurohashi. 2017.
\newblock An empirical comparison of domain adaptation methods for neural
  machine translation.
\newblock In \emph{Proceedings of the 55th Annual Meeting of the Association
  for Computational Linguistics (Volume 2: Short Papers)}, volume~2, pages
  385--391.

\bibitem[{Cohn and Lapata(2007)}]{cohn2007machine}
Trevor Cohn and Mirella Lapata. 2007.
\newblock Machine translation by triangulation: Making effective use of
  multi-parallel corpora.
\newblock In \emph{Proceedings of the 45th Annual Meeting of the Association of
  Computational Linguistics}, pages 728--735.

\bibitem[{Currey et~al.(2017)Currey, Barone, and Heafield}]{currey2017copied}
Anna Currey, Antonio Valerio~Miceli Barone, and Kenneth Heafield. 2017.
\newblock Copied monolingual data improves low-resource neural machine
  translation.
\newblock In \emph{Proceedings of the Second Conference on Machine
  Translation}, pages 148--156.

\bibitem[{De~Gispert and Marino(2006)}]{degispert2006catalan}
Adri{\`a} De~Gispert and Jose~B Marino. 2006.
\newblock Catalan-english statistical machine translation without parallel
  corpus: bridging through spanish.
\newblock In \emph{Proc. of 5th International Conference on Language Resources
  and Evaluation (LREC)}, pages 65--68. Citeseer.

\bibitem[{Gu et~al.(2018)Gu, Hassan, Devlin, and Li}]{gu2018universal}
Jiatao Gu, Hany Hassan, Jacob Devlin, and Victor~OK Li. 2018.
\newblock Universal neural machine translation for extremely low resource
  languages.
\newblock In \emph{Proceedings of the 2018 Conference of the North American
  Chapter of the Association for Computational Linguistics: Human Language
  Technologies, Volume 1 (Long Papers)}, volume~1, pages 344--354.

\bibitem[{Guzm{\'a}n et~al.(2019)Guzm{\'a}n, Chen, Ott, Pino, Lample, Koehn,
  Chaudhary, and Ranzato}]{guzman2019two}
Francisco Guzm{\'a}n, Peng-Jen Chen, Myle Ott, Juan Pino, Guillaume Lample,
  Philipp Koehn, Vishrav Chaudhary, and Marc'Aurelio Ranzato. 2019.
\newblock Two new evaluation datasets for low-resource machine translation:
  Nepali-\uppercase{E}nglish and \uppercase{S}inhala-\uppercase{E}nglish.
\newblock \emph{arXiv preprint arXiv:1902.01382}.

\bibitem[{Hassan et~al.(2017)Hassan, Elaraby, and Tawfik}]{hassan2017synthetic}
Hany Hassan, Mostafa Elaraby, and Ahmed Tawfik. 2017.
\newblock Synthetic data for neural machine translation of spoken-dialects.
\newblock \emph{arXiv preprint arXiv:1707.00079}.

\bibitem[{Johnson et~al.(2017)Johnson, Schuster, Le, Krikun, Wu, Chen, Thorat,
  Vi{\'e}gas, Wattenberg, Corrado et~al.}]{johnson2017google}
Melvin Johnson, Mike Schuster, Quoc~V Le, Maxim Krikun, Yonghui Wu, Zhifeng
  Chen, Nikhil Thorat, Fernanda Vi{\'e}gas, Martin Wattenberg, Greg Corrado,
  et~al. 2017.
\newblock Google’s multilingual neural machine translation system: Enabling
  zero-shot translation.
\newblock \emph{Transactions of the Association for Computational Linguistics},
  5:339--351.

\bibitem[{Klein et~al.(2017)Klein, Kim, Deng, Senellart, and Rush}]{opennmt}
Guillaume Klein, Yoon Kim, Yuntian Deng, Jean Senellart, and Alexander Rush.
  2017.
\newblock Opennmt: Open-source toolkit for neural machine translation.
\newblock In \emph{Proceedings of 55th Annual Meeting of the Association for
  Computational Linguistics, System Demonstrations}, pages 67--72.

\bibitem[{Koehn and Knowles(2017)}]{koehn2017six}
Philipp Koehn and Rebecca Knowles. 2017.
\newblock Six challenges for neural machine translation.
\newblock In \emph{Proceedings of the First Workshop on Neural Machine
  Translation}, pages 28--39.

\bibitem[{Lakew et~al.(2018)Lakew, Lotito, Negri, Turchi, and
  Federico}]{lakew2018improving}
Surafel~M Lakew, Quintino~F Lotito, Matteo Negri, Marco Turchi, and Marcello
  Federico. 2018.
\newblock Improving zero-shot translation of low-resource languages.
\newblock \emph{arXiv preprint arXiv:1811.01389}.

\bibitem[{Lample et~al.(2018{\natexlab{a}})Lample, Conneau, Denoyer, and
  Ranzato}]{lample2017unsupervised}
Guillaume Lample, Alexis Conneau, Ludovic Denoyer, and Marc'Aurelio Ranzato.
  2018{\natexlab{a}}.
\newblock Unsupervised machine translation using monolingual corpora only.
\newblock In \emph{International Conference on Learning Representations}.

\bibitem[{Lample et~al.(2018{\natexlab{b}})Lample, Conneau, Ranzato, Denoyer,
  and Jégou}]{conneau2017word}
Guillaume Lample, Alexis Conneau, Marc'Aurelio Ranzato, Ludovic Denoyer, and
  Hervé Jégou. 2018{\natexlab{b}}.
\newblock Word translation without parallel data.
\newblock In \emph{International Conference on Learning Representations}.

\bibitem[{Lample et~al.(2018{\natexlab{c}})Lample, Ott, Conneau, Denoyer
  et~al.}]{lample2018phrase}
Guillaume Lample, Myle Ott, Alexis Conneau, Ludovic Denoyer, et~al.
  2018{\natexlab{c}}.
\newblock Phrase-based \& neural unsupervised machine translation.
\newblock In \emph{Proceedings of the 2018 Conference on Empirical Methods in
  Natural Language Processing}, pages 5039--5049.

\bibitem[{Levinboim and Chiang(2015)}]{levinboim-chiang:2015:EMNLP}
Tomer Levinboim and David Chiang. 2015.
\newblock Supervised phrase table triangulation with neural word embeddings for
  low-resource languages.
\newblock In \emph{Proceedings of the 2015 Conference on Empirical Methods in
  Natural Language Processing}, pages 1079--1083, Lisbon, Portugal. Association
  for Computational Linguistics.

\bibitem[{Lin et~al.(2019)Lin, Chen, Lee, Li, Zhang, Xia, Rijhwani, He, Zhang,
  Ma, Anastasopoulos, Littell, and Neubig}]{lin19acl}
Yu-Hsiang Lin, Chian-Yu Chen, Jean Lee, Zirui Li, Yuyan Zhang, Mengzhou Xia,
  Shruti Rijhwani, Junxian He, Zhisong Zhang, Xuezhe Ma, Antonios
  Anastasopoulos, Patrick Littell, and Graham Neubig. 2019.
\newblock Choosing transfer languages for cross-lingual learning.
\newblock In \emph{The 57th Annual Meeting of the Association for Computational
  Linguistics (ACL)}, Florence, Italy.

\bibitem[{Litschko et~al.(2018)Litschko, Glava{\v{s}}, Ponzetto, and
  Vuli{\'c}}]{litschko2018unsupervised}
Robert Litschko, Goran Glava{\v{s}}, Simone~Paolo Ponzetto, and Ivan Vuli{\'c}.
  2018.
\newblock Unsupervised cross-lingual information retrieval using monolingual
  data only.
\newblock \emph{arXiv preprint arXiv:1805.00879}.

\bibitem[{Mikolov et~al.(2013)Mikolov, Le, and
  Sutskever}]{mikolov2013exploiting}
Tomas Mikolov, Quoc~V Le, and Ilya Sutskever. 2013.
\newblock Exploiting similarities among languages for machine translation.
\newblock \emph{arXiv preprint arXiv:1309.4168}.

\bibitem[{Neubig and Hu(2018)}]{neubig18emnlp}
Graham Neubig and Junjie Hu. 2018.
\newblock Rapid adaptation of neural machine translation to new languages.
\newblock In \emph{Conference on Empirical Methods in Natural Language
  Processing (EMNLP)}, Brussels, Belgium.

\bibitem[{Nguyen and Chiang(2017)}]{nguyen2017transfer}
Toan~Q. Nguyen and David Chiang. 2017.
\newblock Transfer learning across low-resource, related languages for neural
  machine translation.
\newblock In \emph{Proc. IJCNLP}, volume~2, pages 296--301.

\bibitem[{Nishimura et~al.(2018)Nishimura, Sudoh, Neubig, and
  Nakamura}]{nishimura2018multi-b}
Yuta Nishimura, Katsuhito Sudoh, Graham Neubig, and Satoshi Nakamura. 2018.
\newblock Multi-source neural machine translation with data augmentation.
\newblock In \emph{Proceedings of the First Workshop on Neural Machine
  Translation}.

\bibitem[{Qi et~al.(2018)Qi, Sachan, Felix, Padmanabhan, and
  Neubig}]{qi2018and}
Ye~Qi, Devendra Sachan, Matthieu Felix, Sarguna Padmanabhan, and Graham Neubig.
  2018.
\newblock When and why are pre-trained word embeddings useful for neural
  machine translation?
\newblock In \emph{Proceedings of the 2018 Conference of the North American
  Chapter of the Association for Computational Linguistics: Human Language
  Technologies, Volume 2 (Short Papers)}, volume~2, pages 529--535.

\bibitem[{Ren et~al.(2019)Ren, Zhang, Liu, Zhou, and Ma}]{ren2019unsupervised}
Shuo Ren, Zhirui Zhang, Shujie Liu, Ming Zhou, and Shuai Ma. 2019.
\newblock Unsupervised neural machine translation with \uppercase{SMT} as
  posterior regularization.
\newblock \emph{arXiv preprint arXiv:1901.04112}.

\bibitem[{Sch{\"o}nemann(1966)}]{schonemann1966generalized}
Peter~H Sch{\"o}nemann. 1966.
\newblock A generalized solution of the orthogonal procrustes problem.
\newblock \emph{Psychometrika}, 31(1):1--10.

\bibitem[{Sennrich et~al.(2016)Sennrich, Haddow, and
  Birch}]{sennrich2015improving}
Rico Sennrich, Barry Haddow, and Alexandra Birch. 2016.
\newblock Improving neural machine translation models with monolingual data.
\newblock In \emph{Proceedings of the 54th Annual Meeting of the Association
  for Computational Linguistics (Volume 1: Long Papers)}, volume~1, pages
  86--96.

\bibitem[{Subramanian et~al.(2019)Subramanian, Lample, Smith, Denoyer, Ranzato,
  and Boureau}]{lample2018styletransfer}
Sandeep Subramanian, Guillaume Lample, Eric~Michael Smith, Ludovic Denoyer,
  Marc'Aurelio Ranzato, and Y{-}Lan Boureau. 2019.
\newblock Multiple-attribute text style.
\newblock In \emph{International Conference on Learning Representations}.

\bibitem[{Vaswani et~al.(2017)Vaswani, Shazeer, Parmar, Uszkoreit, Jones,
  Gomez, Kaiser, and Polosukhin}]{vaswani2017attention}
Ashish Vaswani, Noam Shazeer, Niki Parmar, Jakob Uszkoreit, Llion Jones,
  Aidan~N Gomez, {\L}ukasz Kaiser, and Illia Polosukhin. 2017.
\newblock Attention is all you need.
\newblock In \emph{Advances in Neural Information Processing Systems}, pages
  5998--6008.

\bibitem[{Wang et~al.(2019)Wang, Pham, Arthur, and
  Neubig}]{wang2018multilingual}
Xinyi Wang, Hieu Pham, Philip Arthur, and Graham Neubig. 2019.
\newblock Multilingual neural machine translation with soft decoupled encoding.
\newblock In \emph{International Conference on Learning Representations}.

\bibitem[{Xie et~al.(2018)Xie, Yang, Neubig, Smith, and
  Carbonell}]{xie2018neural}
Jiateng Xie, Zhilin Yang, Graham Neubig, Noah~A Smith, and Jaime Carbonell.
  2018.
\newblock Neural cross-lingual named entity recognition with minimal resources.
\newblock In \emph{Proceedings of the 2018 Conference on Empirical Methods in
  Natural Language Processing}, pages 369--379.

\bibitem[{Xing et~al.(2015)Xing, Wang, Liu, and Lin}]{xing-EtAl:2015:NAACL-HLT}
Chao Xing, Dong Wang, Chao Liu, and Yiye Lin. 2015.
\newblock Normalized word embedding and orthogonal transform for bilingual word
  translation.
\newblock In \emph{Proceedings of the 2015 Conference of the North American
  Chapter of the Association for Computational Linguistics: Human Language
  Technologies}, pages 1006--1011, Denver, Colorado. Association for
  Computational Linguistics.

\bibitem[{Zhang et~al.(2017)Zhang, Liu, Luan, and Sun}]{zhang2017earth}
Meng Zhang, Yang Liu, Huanbo Luan, and Maosong Sun. 2017.
\newblock Earth mover’s distance minimization for unsupervised bilingual
  lexicon induction.
\newblock In \emph{Proceedings of the 2017 Conference on Empirical Methods in
  Natural Language Processing}, pages 1934--1945.

\bibitem[{Zhang et~al.(2018)Zhang, Ren, Liu, Wang, Chen, Li, Zhou, and
  Chen}]{zhang2018style}
Zhirui Zhang, Shuo Ren, Shujie Liu, Jianyong Wang, Peng Chen, Mu~Li, Ming Zhou,
  and Enhong Chen. 2018.
\newblock Style transfer as unsupervised machine translation.
\newblock \emph{arXiv preprint arXiv:1808.07894}.

\bibitem[{Zoph et~al.(2016)Zoph, Yuret, May, and Knight}]{zoph2016transfer}
Barret Zoph, Deniz Yuret, Jonathan May, and Kevin Knight. 2016.
\newblock Transfer learning for low-resource neural machine translation.
\newblock In \emph{Proceedings of the 2016 Conference on Empirical Methods in
  Natural Language Processing}, pages 1568--1575.

\end{thebibliography}
\bibliographystyle{acl_natbib}

\end{document}